\documentclass{article} 
\usepackage{iclr2027_conference,times}
\iclrfinalcopy 

\usepackage{amsmath,amsfonts,bm}

\def\eqref#1{equation~\ref{#1}}

\def\1{\bm{1}}

\DeclareMathAlphabet{\mathsfit}{\encodingdefault}{\sfdefault}{m}{sl}
\SetMathAlphabet{\mathsfit}{bold}{\encodingdefault}{\sfdefault}{bx}{n}

\usepackage{hyperref}
\usepackage{url}
\usepackage{graphicx}
\usepackage{booktabs}
\usepackage{multirow}
\usepackage{authblk}

\title{BitNest: Bit-Nested Speculative Decoding for Memory-Efficient LLM Inference Acceleration}

\author[1]{Chence~Yang}
\author[1]{Ningxi~Cheng}
\author[2]{Arash~Akbari}
\author[1]{Qitao~Tan}
\author[1]{Qingchan~Zhu}
\author[1]{Ci~Zhang}
\author[3,2]{Changdi~Yang}
\author[2]{Yanzhi~Wang}
\author[1]{Wei~Niu}
\author[4]{Jinhui~Wang}
\author[1]{Jin~Lu}
\author[1]{Geng~Yuan}

\affil[1]{University of Georgia}
\affil[2]{Northeastern University}
\affil[3]{XPeng Motors}
\affil[4]{The University of Alabama}
\begin{document}

\maketitle

\begin{abstract}
Speculative decoding accelerates autoregressive generation by using a lightweight draft to propose multiple tokens for parallel verification. However, existing methods often require an additional draft model or weight representation, introducing non-negligible memory overhead on resource-constrained devices. Self-speculative approaches reduce this overhead, yet still face trade-offs between draft quality, target quality, and storage efficiency. We propose \textbf{BitNest}, a bit-nested speculative decoding framework that embeds a low-precision draft directly into the higher-precision target representation. Instead of deriving a draft from a predefined target, BitNest first constructs a strong low-precision base and then recovers the higher-precision target through residual refinement, enabling both models to share a single physical weight representation. BitNest further extends this progressive-precision design to the KV cache for long-context inference. Across multiple 7B--8B edge-friendly LLMs and diverse workloads, BitNest achieves an average speculative acceptance rate of \textbf{95.2\%} while closely preserving higher-precision model quality, and delivers \textbf{1.48--1.61$\times$} end-to-end speedup over FP16 autoregressive decoding. On the LLaMA models supported by all representative self-speculative baselines, BitNest also achieves consistently competitive or higher decoding speedup.

\end{abstract}

\section{Introduction}

Despite the rapid advances in large language models (LLMs), autoregressive generation remains constrained by high computational cost, substantial model and KV-cache memory footprint, and intensive memory-bandwidth demand. These limitations become particularly pronounced in resource-constrained deployment settings, such as single-GPU systems, consumer-grade GPUs, and edge devices. In addition to model compression techniques such as quantization and pruning, speculative decoding provides an orthogonal approach to accelerating inference: a lightweight draft model proposes multiple candidate tokens, which are then verified in parallel by the target model, thereby reducing the number of expensive token-by-token target executions. To achieve meaningful speedup, however, the draft must be sufficiently inexpensive while remaining well aligned with the target, so that a high acceptance rate can be maintained.

A straightforward way to construct the draft is to use a smaller language model, but a larger capability gap from the target typically leads to lower acceptance. Recent approaches such as EAGLE \cite{eagle} and DFlash \cite{Dflash} improve drafting quality or efficiency through more sophisticated prediction mechanisms, yet still require additional draft networks, parameters, or dedicated representations. This extra draft model storage is particularly costly on memory-constrained devices, where it competes directly with the target model and KV cache for a limited memory budget. For example, under FP16 inference, a 7B model alone can already approach
the memory capacity of a 16 GB device, leaving little room for the KV cache
and runtime buffers and potentially failing to support the desired context
length. Under such tight memory constraints, even a substantially smaller
draft introduces non-negligible deployment overhead.

Self-speculative decoding avoids maintaining an independent draft model, but existing designs still expose a trade-off between drafting efficiency, target quality, and storage overhead. Layer-skipping \cite{layerskip} approaches reduce drafting cost by using a cheaper execution path, but couple drafting efficiency with prediction quality, while quantization-based methods either constrain the target to a shared low-bit representation, as in QSpec \cite{Qspec}, or retain an additional low-precision draft representation, as in QuantSpec \cite{QuantSpec}. This leads to a natural question: can a single model representation support both a lightweight draft and a higher-quality target without additional memory overhead, while preserving sufficient  draft--target agreement for high acceptance? Quantization provides a natural way to explore this possibility, since different precision levels can represent the same model
at different memory and computation costs. This motivates a more concrete formulation: \textit{can a high-precision representation be constructed to explicitly contain a strong low-precision model as an accessible subset?}

Realizing such a nested precision hierarchy, however, is non-trivial. First, the low-precision representation must be sufficiently accurate to serve as a strong speculative draft and maintain high draft--target agreement. Second, the higher-precision target must be recovered by adding information on top of this fixed base; independently quantizing the higher-precision model would generally change the low-bit solution and break the nesting relationship. These requirements motivate a base-first construction that preserves the draft while refining only the missing information.

Building on this construction, we propose BitNest, a new self-speculative decoding paradigm that we call \emph{Bit-Nested Speculative Decoding}, in which the low-precision draft is physically nested within the higher-precision target representation rather than maintained as a separate model. BitNest first constructs a strong low-precision base and then adds refinement bits without modifying that base. During decoding, the draft accesses only the base, while the target additionally accesses the refinement to recover the higher-precision representation. This allows a single physical model representation to support both a lightweight low-precision draft view and a higher-precision target view without maintaining multiple model copies. We further develop and extend our proposed nesting principle to support bit-nested KV cache for long-context decoding acceleration.

\begin{figure}[th]
    \centering
    \includegraphics[draft=false,width=\linewidth]{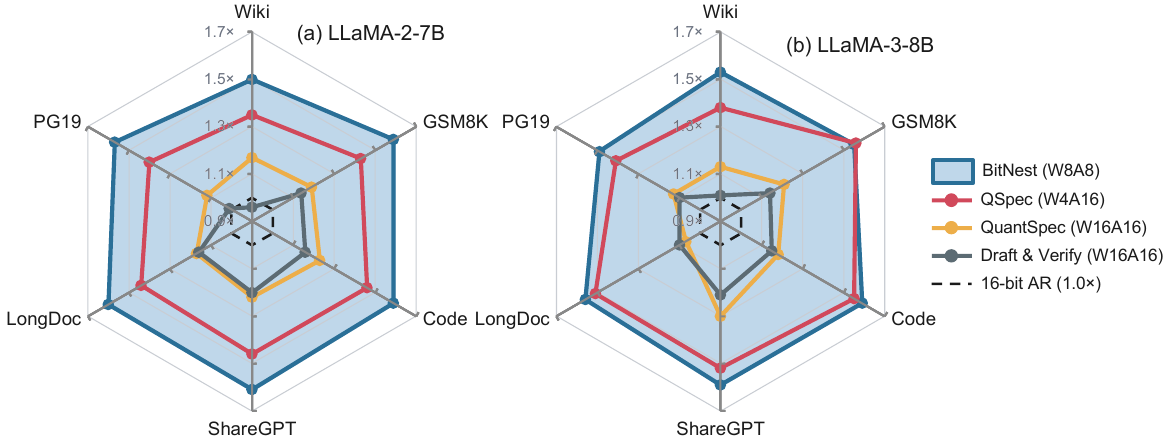}
    \caption{End-to-end decoding speedup across six workloads on LLaMA-2-7B and LLaMA-3-8B. All results are normalized to the corresponding 16-bit autoregressive baseline (1.0$\times$).}
    \label{fig:speedup-radar}
\end{figure}

In our implementation, we instantiate BitNest with the commonly used W4/W8 precision pair, using a W4 base for drafting and a refined W8 representation for verification. We evaluate this realization on LLaMA-2-7B \cite{llama2}, LLaMA-3-8B \cite{llama3}, Qwen2-7B \cite{qwen2}, and Qwen2.5-7B \cite{qwen25} across diverse workloads, including language modeling \cite{wikitext}, mathematical reasoning \cite{gsm8k}, code generation \cite{humaneval} and \cite{mbpp}, conversational data \cite{sharegpt}, and long-context tasks \cite{pg19,longbench,govreport}. BitNest closely preserves the quality of the higher-precision target while maintaining strong draft--target agreement, with an average speculative acceptance rate of 95.2\%. As shown in Figure~\ref{fig:speedup-radar}, BitNest delivers consistently strong end-to-end acceleration across six workloads on both LLaMA-2-7B and LLaMA-3-8B, and achieves 1.48--1.61$\times$ speedup over 16-bit autoregressive decoding across the evaluated models. On a Jetson Orin NX 16GB edge configuration, BitNest further achieves approximately $1.5\times$ decoding speedup on LLaMA-2-7B-32K while extending the maximum executable context from 6K with W8A8 autoregressive decoding to 10K.

Our contributions are as follows:
\begin{itemize}
    \item We introduce Bit-Nested Speculative Decoding, a new self-speculative decoding paradigm in which the draft and target are explicitly co-encoded as nested bit planes within a single physical weight representation.
    \item We develop a base-first construction that optimizes a strong W4 draft and then quantizes its residual error to form a higher-precision W8 target without changing the draft. This preserves the nesting relationship while avoiding the quality limit of a W4 target.
    \item We design dual-plane weight storage so that drafting reads only the base plane and verification reads both planes. We extend the same access principle to a nested KV cache for long-context decoding.
    \item Across four 7B–8B language models and six workloads, BitNest closely preserves higher-precision target quality while achieving 1.48–1.61× decoding speedup over 16-bit autoregressive inference. Edge-device experiments further demonstrate reduced memory use and decode energy.
\end{itemize}

\section{Related Work}

\subsection{Speculative Decoding}

Speculative decoding accelerates autoregressive generation by using a lightweight drafter to propose multiple future tokens for parallel verification by the target model \cite{Spec1,Spec2}. Subsequent work improves candidate generation through multi-token prediction \cite{medusa}, feature-space drafting \cite{eagle,eagle2,eagle3}, and parallel diffusion-based drafting. Recent diffusion-based approaches include TiDAR \cite{tidar}, Your LLM Knows the Future \cite{llmfuture}, DiffuSpec \cite{diffuspec}, SpecDiff-2 \cite{specdiff2}, and DFlash \cite{Dflash}, with DFlash using a lightweight block-diffusion drafter to generate candidate blocks in parallel. Despite increasingly effective drafting mechanisms, these methods still face the trade-off between drafting cost and draft--target agreement, while auxiliary drafters introduce additional parameters or representations that increase deployment and memory overhead.

\subsection{Self-Speculative Decoding}

Self-speculative decoding avoids a separately deployed draft model by deriving a cheaper execution path from the target itself. Layer-skipping and early-exit approaches, including Draft \& Verify \cite{DriftVerify}, LayerSkip \cite{layerskip}, Kangaroo \cite{kangaroo}, and SWIFT \cite{swift}, reduce drafting computation but couple efficiency with the quality of the reduced execution path. Quantization-based methods provide another direction: QSpec \cite{Qspec} reuses shared low-bit weights across drafting and verification, while QuantSpec \cite{QuantSpec} combines low-precision draft weights with hierarchical KV-cache quantization for long-context inference. However, existing designs either constrain the target through a shared low-bit representation or retain additional low-precision draft weights. BitNest instead targets representation-level nesting, where the draft is directly embedded within the higher-precision target representation.

\subsection{LLM Quantization and Rotation-Based PTQ}

Post-training quantization (PTQ) reduces LLM memory and inference cost by representing weights and activations at lower precision. Existing methods improve low-bit quantization through second-order error compensation, activation-aware scaling, clipping, and equivalent transformations \cite{GPTQ,awq,rptq,QuIP1,smoothquant}. 
More recently, rotation- and transformation-based approaches such as QuaRot \cite{QuaRot}, SpinQuant \cite{SpinQuant}, DuQuant \cite{duquant}, and FlatQuant \cite{flatquant} apply function-preserving transformations to suppress or redistribute outliers in hidden states and activations, yielding equivalent model representations that are more robust to low-precision quantization. These methods primarily optimize individual quantized configurations rather than explicitly organizing multiple precision views into a nested representation. BitNest builds on this transformed representation to construct nested draft and target views within the same physical weight representation.

\section{Methodology}
\label{sec:method}

\begin{figure}[h]
\begin{center}
\includegraphics[draft=false,width=\linewidth]{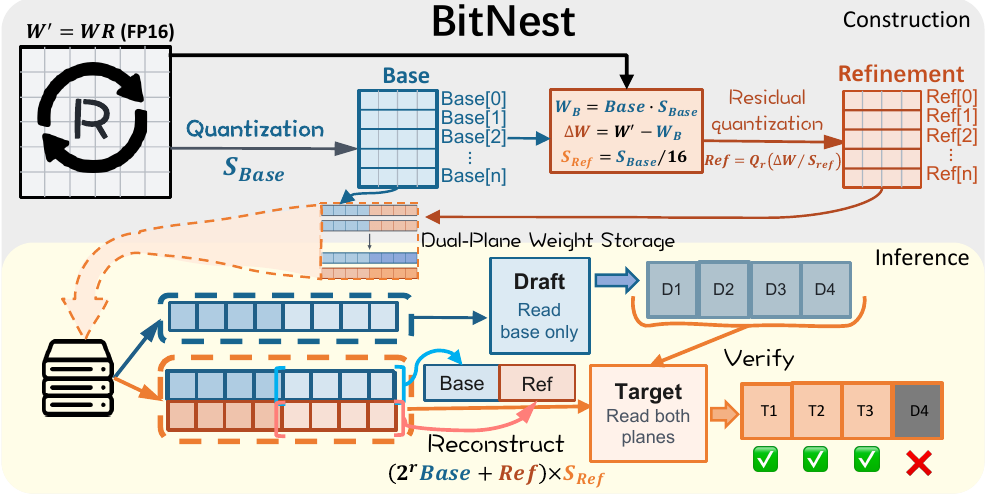}
\end{center}
\caption{Overview of BitNest. BitNest first applies a function-preserving rotation-based transformation to obtain an equivalent model representation, then quantizes the transformed weights into a $b$-bit base and the remaining
error into an $r$-bit refinement. The two are organized using \emph{Dual-Plane Weight Storage}. During speculative decoding, drafting reads only the base, while verification reads both planes to recover the $(b+r)$-bit target. BitNest physically nests the low-precision draft within the higher-precision target without requiring a separate draft model.}
\label{fig:bitnest_method}
\end{figure}\textbf{}
\subsection{Overview}
\label{sec:method_overview}

Existing speculative decoding methods typically rely on either an auxiliary draft model or a cheaper execution path derived from the target. The former introduces additional model storage and potential draft--target mismatch, while the latter often trades drafting cost against prediction quality. BitNest instead represents the draft and target as nested precision views within the same physical model representation.

As illustrated in Figure~\ref{fig:bitnest_method}, BitNest first applies a rotation-based transformation to obtain an equivalent model representation. On the transformed weights, BitNest constructs a $b_W$-bit base plane for drafting and further encodes the remaining weight-quantization residual into an $r_W$-bit refinement plane. Together, the base and refinement define a nested $(b_W+r_W)$-bit target representation:
\begin{equation}
W
\xrightarrow{R}
W'
\xrightarrow{Q_{b_W}}
W_{\mathrm{base}}
\xrightarrow{+\;r_W\text{-bit refinement}}
W_{\mathrm{target}} .
\end{equation}
To enable base-only drafting without fetching refinement bits, we introduce \emph{Dual-Plane Weight Storage}. It packs the base and refinement into separate, independently accessible bit planes, allowing the draft to read only the base and the target to read both.
In this work, we instantiate $b_W=r_W=4$, yielding a W4 draft view and a W8 target view. We focus on the W4/W8 setting because 4-bit and 8-bit integer formats are widely supported by existing quantized inference systems and hardware kernels.

We further extend the nesting principle independently to the KV cache. We denote the KV base and refinement precisions as $b_{\mathrm{KV}}$ and $r_{\mathrm{KV}}$, respectively. In our implementation, $b_{\mathrm{KV}}=r_{\mathrm{KV}}=4$, corresponding to KV4 access during drafting and KV8 access during verification.

\subsection{BitNest Weight Construction}
\label{sec:bitnest}

The BitNest construction is illustrated under the W4/W8 setting adopted in this work. After the function-preserving rotations $R$ are folded into the pretrained weights, the transformed weights are denoted by $W'$. A strong W4 base is
first constructed using GPTQ, producing the integer code $q_4$ and group-wise scale $s_4$, with 
\begin{equation}
    W_{\mathrm{draft}}=q_4s_4 .
\end{equation}
The base is then fixed throughout the subsequent refinement.
The remaining quantization error $\Delta W=W'-W_{\mathrm{draft}}$ is encoded by an additional 4-bit refinement $q_r$. Using a refinement scale $s_{\mathrm{ref}}=s_4/16$, the target representation is reconstructed as
\begin{equation}
    W_{\mathrm{target}}
    = q_4s_4 + q_rs_{\mathrm{ref}}
    = (16q_4+q_r)s_{\mathrm{ref}} .
\end{equation}
Thus, the higher-precision target is recovered without modifying the
optimized W4 base, making the draft representation an exact subset of the
W8 target. The precise refinement quantization rule and signed-integer
boundary handling are provided in Appendix~\ref{app:weight_construction}.

During speculative decoding, the draft stage accesses only the fixed W4 base,
while verification additionally accesses the refinement to reconstruct the
W8 target model. The two stages therefore operate on different precision views of
the same physical representation rather than on separately stored models.

\subsection{Dual-Plane Weight Storage}
\label{sec:efficient_execution}

\begin{figure}[h]

\begin{center}
\includegraphics[draft=false,width=\linewidth]{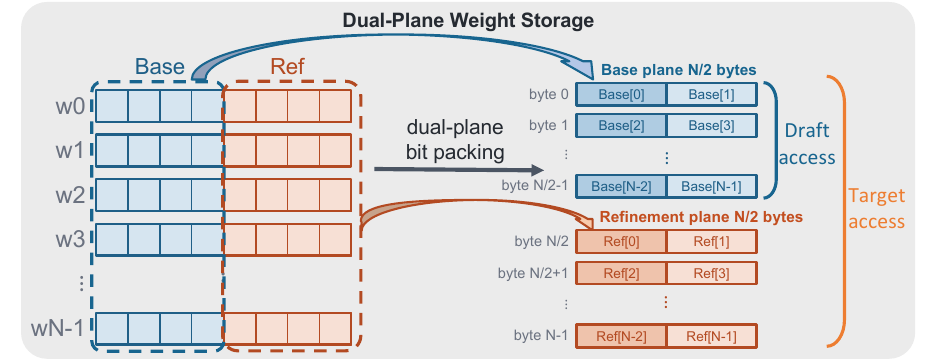}
\end{center}
\caption{Dual-plane weight storage in BitNest. Base and refinement codes are packed into two independent 4-bit streams, each occupying $N/2$ bytes for $N$ weights. The draft accesses only the base plane, while the target accesses both planes to recover the full 8-bit representation.}
\label{fig:dual-plane}
\end{figure}

BitNest provides a W4 draft and a refined W8 target from the same nested representation. However, the logical reduction from W8 to W4 does not automatically translate into lower memory traffic. Modern memory systems are byte-addressable, and an individual 4-bit nibble cannot be independently addressed. If the base and refinement of each weight were stored together in a conventional byte, draft execution would still fetch the refinement bits even though only the W4 base is required. This would largely eliminate the memory-access advantage of the low-precision draft. BitNest therefore stores the base and refinement as two independently packed bit planes, as illustrated in Figure~\ref{fig:dual-plane}.

For $N$ weights, the base and refinement planes each occupy $N/2$ bytes. As illustrated by the inference path in Figure~\ref{fig:bitnest_method}, the draft accesses only the base plane, corresponding to 4 bits/weight of weight-code traffic, while verification accesses both planes to reconstruct the W8 target. This layout allows the draft to avoid fetching refinement bits. The refinement requires no additional scale metadata because $s_{\mathrm{ref}}=s_4/16$ is derived from the base scale.




\subsection{Context-Aware KV Nesting}
\label{sec:kv_nesting}

\begin{figure}[h]

\begin{center}
\includegraphics[draft=false,width=\linewidth]{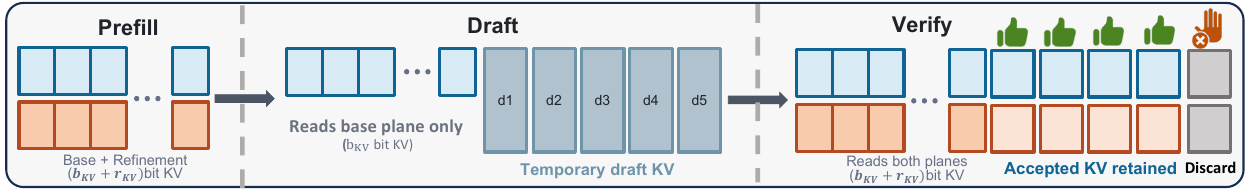}
\end{center}
\caption{KV nesting during speculative decoding.
The draft stage only reads the low-bit base view, while verification reads the full $b_{kv}+r_{kv}$-bit view from the same nested cache. Accepted KV entries are retained while the rejected suffix is discarded.}
\label{fig:kv_nesting}
\end{figure}

As the context grows, KV-cache traffic accounts for an increasing fraction
of decoding cost, reducing the benefit of weight-side savings alone.
As shown in Figure~\ref{fig:long_context}, the benefit of lower-precision
draft KV access becomes increasingly pronounced at longer contexts. BitNest therefore extends the nesting principle to the KV cache: the draft reads only the low-precision base, while the target additionally
accesses the refinement to recover the higher-precision KV states.
As illustrated in Figure~\ref{fig:kv_nesting}, the draft reads only the KV4
base plane, while verification reads both planes to recover the KV8 historical
cache. KV states generated for candidate tokens during drafting are kept
temporarily; during verification, the target recomputes these states, retaining
the accepted entries and discarding the rejected suffix. At short contexts,
BitNest uses KV8 for both drafting and verification to preserve draft--target
agreement, while at long contexts it switches to KV4 drafting and KV8
verification to reduce KV-cache traffic. Detailed KV quantization is provided
in Appendix~\ref{app:kv_quant}.

\section{Experiment}
\label{sec:experiments}
\label{headings}
\subsection{Experiment setting}



We evaluate BitNest on LLaMA-2-7B, LLaMA-3-8B, Qwen2-7B, and Qwen2.5-7B across six workloads: WikiText-2, GSM8K, Code (HumanEval, MBPP), ShareGPT, LongDoc, and PG-19. Long-context evaluation additionally uses LLaMA-2-7B-32K. All main experiments are conducted on a single NVIDIA RTX A6000 GPU. Detailed implementation, baseline, and evaluation settings are provided in Appendix~\ref{app:exp_setup}.

\subsection{Speedup for LLM inference}

\begin{table}[th]
\centering
\caption{End-to-end decoding speedup across six workloads. Speedup is measured over the corresponding 16-bit autoregressive baseline in the same inference engine. N/A denotes configurations not supported by the official implementations for the evaluated Qwen models.}
\label{tab:llm_speedup}

\small
\setlength{\tabcolsep}{3.5pt}
\renewcommand{\arraystretch}{1.08}

\begin{tabular}{lllcccccc}
\toprule
\textbf{Model} & \textbf{Method} & \textbf{Draft $\rightarrow$ Target}
& \textbf{Wiki}
& \textbf{GSM8K}
& \textbf{Code}
& \textbf{ShareGPT}
& \textbf{LongDoc}
& \textbf{PG19} \\
\midrule

\multirow{4}{*}{LLaMA-2-7B}
& QSpec
& W4A4 $\rightarrow$ W4A16
& 1.35$\times$ & 1.43$\times$ & 1.46$\times$
& 1.46$\times$ & 1.44$\times$ & 1.40$\times$ \\

& QuantSpec
& W4A16 $\rightarrow$ W16A16
& 1.17$\times$ & 1.19$\times$ & 1.23$\times$
& 1.22$\times$ & 1.17$\times$ & 1.12$\times$ \\

& Draft \& Verify
& W16A16 $\rightarrow$ W16A16
& 0.96$\times$ & 1.14$\times$ & 1.16$\times$
& 1.20$\times$ & 1.16$\times$ & 1.01$\times$ \\

& \textbf{BitNest}
& \textbf{W4A8 $\rightarrow$ W8A8}
& \textbf{1.50$\times$} & \textbf{1.59$\times$}
& \textbf{1.59$\times$} & \textbf{1.61$\times$}
& \textbf{1.60$\times$} & \textbf{1.57$\times$} \\

\midrule

\multirow{4}{*}{LLaMA-3-8B}
& QSpec
& W4A4 $\rightarrow$ W4A16
& 1.38$\times$ & 1.56$\times$ & 1.55$\times$
& 1.52$\times$ & 1.51$\times$ & 1.41$\times$ \\

& QuantSpec
& W4A16 $\rightarrow$ W16A16
& 1.13$\times$ & 1.21$\times$ & 1.18$\times$
& 1.30$\times$ & 1.07$\times$ & 1.13$\times$ \\

& Draft \& Verify
& W16A16 $\rightarrow$ W16A16
& 1.01$\times$ & 1.14$\times$ & 1.15$\times$
& 1.21$\times$ & 1.10$\times$ & 1.10$\times$ \\

& \textbf{BitNest}
& \textbf{W4A8 $\rightarrow$ W8A8}
& \textbf{1.53$\times$} & \textbf{1.55$\times$}
& \textbf{1.59$\times$} & \textbf{1.59$\times$}
& \textbf{1.56$\times$} & \textbf{1.49$\times$} \\

\midrule

\multirow{4}{*}{Qwen2-7B}
& QSpec
& W4A4 $\rightarrow$ W4A16
& N/A & N/A & N/A & N/A & N/A & N/A \\

& QuantSpec
& W4A16 $\rightarrow$ W16A16
& N/A & N/A & N/A & N/A & N/A & N/A \\

& Draft \& Verify
& W16A16 $\rightarrow$ W16A16
& N/A & N/A & N/A & N/A & N/A & N/A \\

& \textbf{BitNest}
& \textbf{W4A8 $\rightarrow$ W8A8}
& \textbf{1.48$\times$} & \textbf{1.53$\times$}
& \textbf{1.54$\times$} & \textbf{1.55$\times$}
& \textbf{1.53$\times$} & \textbf{1.52$\times$} \\

\midrule

\multirow{4}{*}{Qwen2.5-7B}
& QSpec
& W4A4 $\rightarrow$ W4A16
& N/A & N/A & N/A & N/A & N/A & N/A \\

& QuantSpec
& W4A16 $\rightarrow$ W16A16
& N/A & N/A & N/A & N/A & N/A & N/A \\

& Draft \& Verify
& W16A16 $\rightarrow$ W16A16
& N/A & N/A & N/A & N/A & N/A & N/A \\

& \textbf{BitNest}
& \textbf{W4A8 $\rightarrow$ W8A8}
& \textbf{1.52$\times$} & \textbf{1.58$\times$}
& \textbf{1.57$\times$} & \textbf{1.58$\times$}
& \textbf{1.59$\times$} & \textbf{1.57$\times$} \\

\bottomrule
\end{tabular}
\end{table}

\begin{table}[th]
\centering
\caption{
Model performance across six evaluation domains. GSM8K reports accuracy ($\uparrow$), while all other datasets report perplexity ($\downarrow$). QSpec uses a W4A16 target and is evaluated with its corresponding inference engine.
}
\label{tab:llm_quality}

\small
\setlength{\tabcolsep}{4pt}
\renewcommand{\arraystretch}{1.06}

\begin{tabular}{llcccccc}
\toprule
Model & Method
& WikiText-2$\downarrow$
& GSM8K$\uparrow$
& Code$\downarrow$
& ShareGPT$\downarrow$
& LongDoc$\downarrow$
& PG19$\downarrow$ \\
\midrule

\multirow{4}{*}{LLaMA-2-7B}
& FP16
& 5.4706 & 13.6 & 2.6069 & 2.2206 & 4.3336 & 9.6594 \\
& W8A8
& 5.4762 & 13.4 & 2.6101 & 2.2229 & 4.3374 & 9.6686 \\
& QSpec (W4A16)
& 5.650 & 12.5 & 2.642 & 2.389 & 4.640 & 10.052 \\
& BitNest
& 5.5123 & 13.8 & 2.6123 & 2.2321 & 4.3540 & 9.7616 \\

\midrule

\multirow{4}{*}{LLaMA-3-8B}
& FP16
& 6.1389 & 50.1 & 2.7131 & 2.6462 & 5.2592 & 11.8364 \\
& W8A8
& 6.1617 & 50.0 & 2.7192 & 2.6551 & 5.2755 & 11.8722 \\
& QSpec (W4A16)
& 6.383 & 37.0 & 2.680 & 2.407 & 5.595 & 12.366 \\
& BitNest
& 6.2256 & 49.8 & 2.7327 & 2.6639 & 5.3117 & 11.9476 \\

\midrule

\multirow{3}{*}{Qwen2-7B}
& FP16
& 7.1337 & 79.4 & 2.2112 & 2.3278 & 5.2482 & 13.7653 \\
& W8A8
& 7.1519 & 78.9 & 2.2131 & 2.3319 & 5.2582 & 13.8070 \\
& BitNest
& 7.1759 & 77.6 & 2.2274 & 2.3397 & 5.2781 & 13.8628 \\

\midrule

\multirow{3}{*}{Qwen2.5-7B}
& FP16
& 6.8507 & 82.9 & 2.0979 & 2.3959 & 5.5095 & 12.8805 \\
& W8A8
& 6.8723 & 83.2 & 2.1011 & 2.4003 & 5.5193 & 12.9236 \\
& BitNest
& 6.9062 & 82.0 & 2.1124 & 2.4061 & 5.5384 & 12.9981 \\

\bottomrule
\end{tabular}
\end{table}

Table~\ref{tab:llm_speedup} reports end-to-end decoding speedup across six workloads. BitNest achieves 1.48--1.61$\times$ speedup over the corresponding 16-bit autoregressive baseline across all four models, with consistent gains across workloads. Compared with existing self-speculative methods, BitNest combines a W4A8 draft with a W8A8 target within the same nested representation, providing strong acceleration without maintaining a separate draft-weight copy.

BitNest also closely preserves the quality of the higher-precision target. As shown in Table~\ref{tab:llm_quality}, its performance remains close to FP16 and standalone W8A8 across all evaluated models and workloads. For example, on LLaMA-3-8B, BitNest achieves 49.8\% GSM8K accuracy, compared with 50.0\% for W8A8. Detailed comparisons with lower-precision W4 targets are provided in Appendix~\ref{app:ablation}. BitNest also achieves an average generation-time acceptance rate of 95.2\% across the four models and six workloads, indicating strong agreement between its W4A8 draft and W8A8 target. Detailed per-model and per-workload results are provided in Appendix~\ref{app:acceptance}.

We additionally evaluate BitNest on OpenVLA-7B \cite{openvla}, where it achieves
1.35$\times$ decode speedup over FP16 autoregressive decoding while
preserving task-level performance; detailed results are provided in
Appendix~\ref{app:vla}.

\subsection{Speedup at Long Context Lengths}
\label{sec:long_context}
\begin{figure}[th]
    \centering
    \includegraphics[draft=false,width=\linewidth]{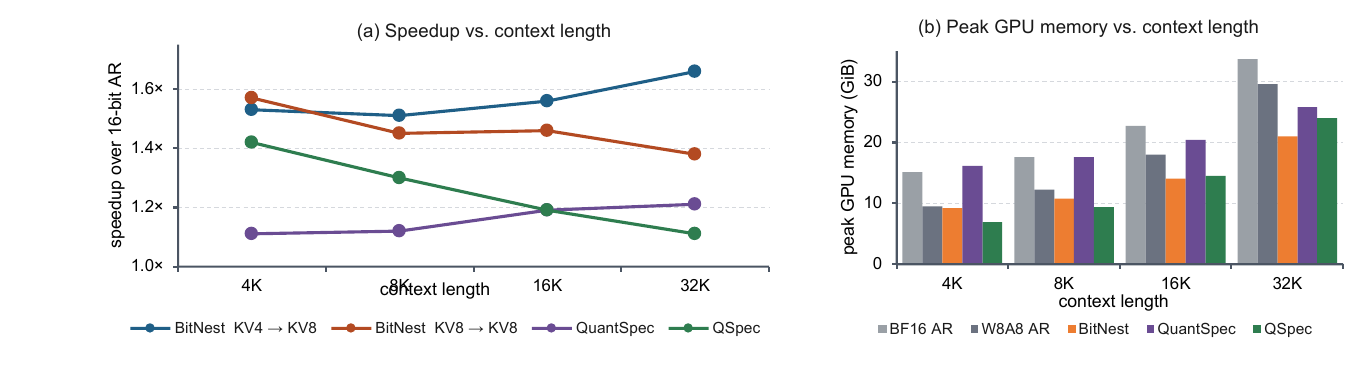}
    \caption{
    Long-context efficiency on LLaMA-2-7B-32K.
    (a) Decoding speedup over the corresponding 16-bit autoregressive
    baseline in the same inference engine. Reading only the KV4 base during
    drafting increasingly benefits BitNest as the context grows.
    (b) Peak GPU memory across context lengths. Weight quantization reduces
    the persistent model footprint, while the nested KV representation
    further limits memory growth at long context lengths.
    }
    \label{fig:long_context}
\end{figure}


As the context length increases, KV-cache access and attention account for an increasingly large fraction of the decoding cost. As shown in Figure~\ref{fig:long_context}(a), when both the draft and target read the full KV8 representation, BitNest speedup decreases from $1.57\times$ at 4K to $1.38\times$ at 32K. In contrast, when the draft reads only the KV4 base plane while the target retains the full KV8 representation, BitNest achieves a $1.66\times$ speedup at 32K. In comparison, QSpec decreases from $1.42\times$ to $1.11\times$, while QuantSpec remains around     $1.11$--$1.21\times$. These results show that BitNest context-aware KV nesting becomes increasingly beneficial as the context length grows.


\subsection{Memory Saving}


\begin{table}[th]
\centering
\caption{Peak GPU memory (GiB) under different context lengths.
Values in parentheses denote the corresponding 16-bit autoregressive
baseline in the same inference engine.}
\label{tab:short_memory}
\small
\setlength{\tabcolsep}{6pt}
\begin{tabular}{llccc}
\toprule
Method & Model & 1K & 2K & 4K \\
\midrule
\multirow{2}{*}{QSpec}
& LLaMA-2-7B & 4.60 (13.42) & 5.35 (14.09) & 6.10 (14.84) \\
& LLaMA-3-8B & 5.90 (15.40) & 6.34 (15.73) & 6.67 (16.40) \\
\midrule
\multirow{2}{*}{QuantSpec}
& LLaMA-2-7B & 13.21 (13.32) & 13.47 (13.82) & 13.93 (14.57) \\
& LLaMA-3-8B & 15.34 (15.33) & 15.64 (15.46) & 16.27 (15.71) \\
\midrule
\multirow{2}{*}{Draft \& Verify}
& LLaMA-2-7B & 14.21 (13.87) & 15.47 (14.94) & 18.33 (18.33) \\
& LLaMA-3-8B & 15.96 (15.96) & 16.82 (16.82) & 20.81 (20.81) \\
\midrule
\multirow{4}{*}{BitNest}
& LLaMA-2-7B & 7.22 (13.32) & 7.57 (13.92) & 8.09 (14.83) \\
& LLaMA-3-8B & 8.96 (15.26) & 9.13 (15.50) & 9.48 (16.01) \\
& Qwen2-7B & 8.61 (14.42) & 8.76 (14.62) & 9.07 (15.02) \\
& Qwen2.5-7B & 8.61 (14.42) & 8.76 (14.62) & 9.07 (15.02) \\
\bottomrule
\end{tabular}
\end{table}

BitNest reduces memory overhead by eliminating a separately stored draft
representation. On LLaMA-2-7B, the nested weights occupy 6.22 GiB,
identical to a standalone W8 model, whereas storing an independent W4
draft together with the INT8 target requires 9.38 GiB. As shown in
Table~\ref{tab:short_memory}, BitNest also reduces end-to-end peak memory across
all evaluated models. Qwen2-7B and Qwen2.5-7B share identical architectures, hence identical memory. QSpec has a smaller footprint at short contexts
because it uses a shared W4 weight representation, but this also limits
its target to W4 precision.

As the context grows, the KV cache accounts for an increasing fraction of
the memory footprint, and the benefit of BitNest becomes more pronounced.
As shown in Figure~\ref{fig:long_context}(b), BitNest achieves the lowest
peak memory among the compared methods at 32K context, using 21.0 GiB
compared with 24.0 GiB for QSpec and 25.1 GiB for QuantSpec.

\subsection{Edge-Device Evaluation}
\label{sec:edge}


\begin{table}[th]
\centering
\caption{
Edge-device decoding performance on NVIDIA Jetson Orin NX.
Energy is averaged over the decode phase.
}
\label{tab:edge}
\small
\begin{tabular}{llccc}
\toprule
Model (Device) & Method & Throughput & Speedup & Energy \\
 & & (tok/s) & & (J/token) \\
\midrule
\multirow{3}{*}{LLaMA-2-7B-32K (NX16)}
& FP16 AR & 6.0 & 1.00$\times$ & 1.52 \\
& W8A8 AR & 7.8 & $\sim$1.30$\times$ & 1.15 \\
& BitNest & 9.0--9.3 & $\sim$1.50$\times$ & \textbf{0.95} \\
\midrule
\multirow{3}{*}{Llama-3.2-3B (NX8)}
& FP16 AR & 10.6 & 1.00$\times$ & 0.768 \\
& W8A8 AR & 14.0 & 1.32$\times$ & 0.54 \\
& BitNest & 15.5--16.6 & 1.46--1.57$\times$ & \textbf{0.46} \\
\bottomrule
\end{tabular}
\end{table}

We further evaluate BitNest on NVIDIA Jetson Orin NX devices with
16GB and 8GB memory. As shown in Table~\ref{tab:edge}, on the 16GB
device, BitNest achieves approximately $1.5\times$ speedup on
LLaMA-2-7B-32K and reduces decode energy from 1.52 to 0.95 J/token
relative to FP16. On the 8GB device, BitNest achieves up to
$1.57\times$ speedup on Llama-3.2-3B while reducing decode energy
from 0.768 to 0.46 J/token. BitNest also extends the maximum executable
context of LLaMA-2-7B-32K from 2K with FP16 to 10K.
Additional edge-device results are provided in the Appendix \ref{app:edge}.
\section{Ablation Study}
\label{sec:ablation}

\subsection{Effect of speculation length}
The speculation length $\gamma$ trades off verification efficiency against draft acceptance. On LLaMA-2-7B-32K, throughput peaks at $\gamma=4$ for both 4K and 31K contexts, reaching 59.3 and 34.3 tok/s, respectively, while larger $\gamma$ further reduces acceptance without improving throughput. Detailed results are provided in Appendix~\ref{app:spec_length}.

\begin{table}[th]
\centering
\caption{
Why higher-precision representations are retained for the target.
Low-precision representations provide effective drafts, while KV8 and
W8 preserve the fidelity required for final generation.
}
\label{tab:target_precision_ablation}
\small
\setlength{\tabcolsep}{4pt}
\renewcommand{\arraystretch}{1.08}

\begin{minipage}[t]{0.49\linewidth}
\centering
\textbf{(a) Why retain KV8 for the target?}

\vspace{2pt}
\begin{tabular}{lcc}
\toprule
Configuration & GSM8K$\uparrow$ & LongBench$\uparrow$ \\
\midrule
KV8 target AR & 82.4 & 41.4 \\
KV8 draft $\rightarrow$ KV8 & 82.2  & 41.2  \\
KV4 draft $\rightarrow$ KV8 & 81.6  & 41.3  \\
KV4 target AR & 80.2 & 41.5 \\
KV4 draft $\rightarrow$ KV4 & 78.4  & 41.4  \\
\bottomrule
\end{tabular}
{\scriptsize Results are measured on Qwen2.5-7B.}
\vspace{2pt}

\end{minipage}
\hfill
\begin{minipage}[t]{0.49\linewidth}
\centering
\textbf{(b) Why retain W8 for the target?}

\vspace{2pt}
\begin{tabular}{lcc}
\toprule
Configuration & WikiText-2$\downarrow$ & GSM8K$\uparrow$ \\
\midrule
W8A8 & 6.162 & 50.0 \\
W4A16 & 6.470 & 41.8 \\
W4A8 & 6.476 & 41.4 \\
BitNest target & 6.223 & 49.8 \\
\bottomrule
\end{tabular}

\vspace{2pt}
{\scriptsize Results are measured on LLaMA-3-8B.}
\end{minipage}
\end{table}

\subsection{Why retain KV8 for the target?}

The fact that KV4 is effective for drafting does not imply that it is
sufficient for the final target. Table~\ref{tab:target_precision_ablation}(a)
shows that reducing the target KV cache from KV8 to KV4 decreases
Qwen2.5 GSM8K accuracy from 81.6\% to 78.4\%. Yet speculative
acceptance remains high at 0.933, showing that high draft--target
agreement alone does not guarantee target fidelity. We therefore use
KV4 on the speculative path while retaining KV8 for verification.
Additional results on model- and task-dependent sensitivity to target
KV precision are provided in Appendix~\ref{app:target_precision}.

\subsection{Why retain W8 for the target?}
The same distinction appears in the weight representation. Although the W4 base provides sufficiently accurate proposals for speculative decoding, using it as the final model incurs a substantial quality loss. As shown in Table \ref{tab:target_precision_ablation}(b), on LLaMA-3-8B, W4A8 achieves 41.4\% GSM8K accuracy compared with 50.0\% for W8A8. Increasing activation precision to W4A16 improves accuracy only slightly to 41.8\%, indicating that the dominant degradation comes from the 4-bit weight representation rather than activation quantization. Reading the refinement plane recovers the BitNest target to 49.8\%. These results support the central nested design: the low-precision base is sufficient for inexpensive drafting, while the refined representation is retained wherever final-output fidelity matters.
Additional ablations on draft KV precision across different attention architectures and more detailed target-precision results are provided in Appendix~\ref{app:draft_kv} and Appendix~\ref{app:target_precision}.

\section{Limitation}
BitNest focuses on self-speculative decoding, where the draft is derived directly from the target representation to minimize additional model storage. This design does not currently exploit auxiliary drafting architectures that can provide greater candidate-generation parallelism. For example, diffusion-based approaches such as DFlash generate candidate blocks in parallel and may achieve stronger acceleration when the additional draft-model overhead is acceptable. Extending BitNest to support such auxiliary drafters, while retaining the memory benefits of nested representations, is an important direction for future work.

\section*{AI use statement}

In this work, we used generative AI tools for implementing experimental methods, executing experiments, and organizing experimental outputs based on the methodology, experimental plan, and datasets specified by the authors. We have not used generative AI tools for developing the theoretical or conceptual framework, proposing research hypotheses or mathematical claims, designing the research methodology or experiments, conducting qualitative data analysis, or interpreting experimental results, and the rest of the required disclosure tasks are not applicable to this work. Additionally, we used generative AI tools for writing and editing experimental code, language editing, improving clarity and conciseness, and refining the organization and presentation of the manuscript. We have reviewed all AI-assisted work. The authors reviewed and tested the AI-assisted implementation and verified the resulting experimental data for correctness and consistency. We take responsibility for the final content of this work, including text, claims or artifacts produced with the aid of generative AI.

\bibliography{iclr2027_conference}
\bibliographystyle{iclr2027_conference}

\appendix
\section{Appendix}
\subsection{Detailed Bit-Nested Weight Construction}
\label{app:weight_construction}

This section provides the detailed construction of the W4/W8 instantiation used in BitNest. The function-preserving $R_1/R_2$ transformations are folded into the pretrained weights $W$, yielding the transformed weights
\begin{equation}
    W' = \mathrm{Fuse}(W;R_1,R_2).
\end{equation}

\paragraph{W4 base construction.}
The low-precision base is constructed using GPTQ with group size 128:
\begin{equation}
    (q_4,s_4)
    =
    \mathrm{Quantize}_{4,g=128}(W'),
    \qquad
    q_4\in[-8,7],
\end{equation}
where $s_4$ denotes the group-wise quantization scale parameter. The corresponding draft weights are reconstructed as
\begin{equation}
    W_{\mathrm{draft}} = q_4s_4 .
\end{equation}
After quantization, both $q_4$ and $s_4$ are fixed so that the subsequent
refinement does not alter the W4 draft representation.

\paragraph{Fixed-base refinement.}
Given the fixed W4 base, the remaining quantization error is
\begin{equation}
    \Delta W = W' - W_{\mathrm{draft}} .
\end{equation}
The refinement scale is defined as
\begin{equation}
    s_{\mathrm{ref}} = \frac{s_4}{16},
\end{equation}
and the corresponding 4-bit refinement code is
\begin{equation}
q_r =
\begin{cases}
\operatorname{clip}
\left(
    \operatorname{round}
    \left(
        \dfrac{\Delta W}{s_{\mathrm{ref}}}
    \right),
    0,7
\right),
& q_4=-8, \\[6pt]
\operatorname{clip}
\left(
    \operatorname{round}
    \left(
        \dfrac{\Delta W}{s_{\mathrm{ref}}}
    \right),
    -8,7
\right),
& q_4>-8 .
\end{cases}
\end{equation}

The asymmetric constraint for $q_4=-8$ prevents the composed integer
code from crossing the lower bound of the signed 8-bit range.

Combining the fixed base and refinement gives
\begin{equation}
\begin{aligned}
    W_{\mathrm{target}}
    &= q_4s_4 + q_rs_{\mathrm{ref}} \\
    &= (16q_4+q_r)s_{\mathrm{ref}} .
\end{aligned}
\end{equation}
Thus, the 8-bit target is recovered without modifying the
optimized W4 base, preserving the W4 draft as an exact subset of the
resulting W8 representation.

\subsection{Detailed KV-Cache Nesting}
\label{app:kv_quant}

All KV-cache constructions follow the KV4/KV8 setting used in our experiments. BitNest performs online per-token, per-KV-head symmetric round-to-nearest (RTN) quantization when a KV entry is written to the cache. For a key or value vector $x\in\mathbb{R}^{d}$, where\ $d=128$ is the head dimension, the scale parameter is
\begin{equation}
    s=\frac{\max_i |x_i|}{7},
    \qquad
    q_4=
    \operatorname{clip}
    \left(
        \left\lfloor \frac{x}{s}\right\rceil,
        -8,7
    \right).
\end{equation}
The corresponding KV4 representation used for drafting is
\begin{equation}
    \hat{x}_{\mathrm{draft}}=q_4s.
\end{equation}
The remaining quantization error is encoded by an additional 4-bit
refinement with scale $s/16$:
\begin{equation}
    q_r=
    \operatorname{clip}
    \left(
        \left\lfloor
        \frac{x-q_4s}{s/16}
        \right\rceil,
        -8,7
    \right),
\end{equation}
giving the KV8 target representation
\begin{equation}
    \hat{x}_{\mathrm{target}}
    =
    \frac{(16q_4+q_r)s}{16}.
\end{equation}
Thus, each KV entry is stored once as nested base and refinement planes
and can be accessed either as KV4 or KV8.

During drafting, BitNest reads only the KV4 base for the historical
context, while verification reads both planes to recover the KV8 cache.
KV states generated for proposed tokens are temporary during drafting
and are recomputed during verification; accepted entries are retained
and the rejected suffix is discarded.

At short contexts, BitNest uses KV8 for both drafting and verification,
while at long contexts it uses KV4 drafting and KV8 verification to
reduce KV-cache traffic. 
BitNest also supports a Recent-$N$ policy as an intermediate operating
point. Under this policy, the most recent $N$ KV entries are kept at the
higher-precision view for drafting, while older entries are accessed through
the low-precision base. This provides a tunable trade-off between draft
fidelity and KV-cache traffic without changing the underlying nested storage.

\subsection{Experimental Setup}
\label{app:exp_setup}

\paragraph{Models and BitNest configuration.}
Experiments are conducted on LLaMA-2-7B, LLaMA-3-8B, Qwen2-7B,
and Qwen2.5-7B, covering both MHA and GQA architectures.
Long-context experiments additionally use LLaMA-2-7B-32K,
following the model setting used by QuantSpec.
BitNest applies SpinQuant rotations \cite{SpinQuant}, followed by GPTQ W4 quantization \cite{GPTQ}
with a group size of 128 to construct the base plane and a closed-form
signed 4-bit residual refinement for the target. Activations use per-token
INT8 quantization, while the KV cache follows the base-plus-refinement
representation described in Section~3.4.

\paragraph{Hardware and implementation.}
All main LLM experiments are conducted on a single NVIDIA RTX A6000 GPU
with batch size 1. BitNest is implemented with custom Triton kernels for
dual-plane GEMM and flash decoding.

\paragraph{Baselines.}
BitNest is compared against QSpec \cite{Qspec}, QuantSpec \cite{QuantSpec}, and Draft \& Verify \cite{DriftVerify} using their official implementations and recommended configurations. The draft and target precisions therefore differ across methods, reflecting
their respective designs rather than a manually unified precision
setting: QSpec uses W4A4$\rightarrow$W4A16, QuantSpec uses
W4A16$\rightarrow$W16A16, Draft \& Verify uses
W16A16$\rightarrow$W16A16, and BitNest uses
W4A8$\rightarrow$W8A8. We preserve these native configurations to avoid
altering the intended operating point of each method. All reported
speedups are measured relative to the corresponding 16-bit
autoregressive baseline in the same inference engine.

\paragraph{Workloads and evaluation protocol.}
Evaluation covers six workloads: WikiText-2, GSM8K, Code, ShareGPT,
LongBench GovReport, and PG-19. All
standard-context evaluations use a context length of 2,048 tokens. For the Code workload, we concatenate
the HumanEval test set (prompt and canonical solution; 164 problems)
with the first 300 problems from the MBPP test set (problem description
and reference solution). The resulting text is tokenized and split into
2,048-token segments, and perplexity is reported over the first 20
segments. Long-context experiments use PG-19 prefixes ranging from 4K
to 32K tokens. Unless otherwise specified, decoding is greedy and
generates 256 new tokens; QuantSpec follows its official 64-token
setting. We report target quality, decoding throughput, speedup, peak
GPU memory, and speculative acceptance rate.

\subsection{Experiment on Vision-Language-Action Model}
\label{app:vla}
We further evaluate BitNest on OpenVLA-7B to examine its potential
applicability beyond text generation. OpenVLA represents each action
chunk using seven autoregressively generated action tokens, making it
compatible with speculative decoding. In contrast, many recent VLA
models \cite{openvla-oft} and \cite{pi0.5} employ flow-matching or related action heads that generate an
action chunk in parallel rather than through autoregressive token
generation. Therefore, this experiment is intended as an exploratory
evaluation of BitNest in the VLA domain, rather than a comparison with
recent VLA architectures. We use the LIBERO-fine-tuned checkpoint and
evaluate task success on LIBERO \cite{libero} Spatial.

\begin{table}[th]
\centering
\caption{OpenVLA-7B performance on LIBERO Spatial. Success rate is
measured using the same evaluation protocol for all precision
configurations.}
\label{tab:openvla_quality}
\small
\setlength{\tabcolsep}{9pt}
\begin{tabular}{lcc}
\toprule
Method & Precision & Success Rate (\%) \\
\midrule
FP16        & FP16   & 86 \\
W8A8        & W8A8   & 87 \\
W4A8        & W4A8   & 80 \\
\textbf{BitNest} & W4A8 $\rightarrow$ W8A8 & \textbf{89} \\
\bottomrule
\end{tabular}
\end{table}
\paragraph{Task performance.}
As shown in Table~\ref{tab:openvla_quality}, directly reducing OpenVLA to W4A8 decreases the success rate from 86\% with FP16 to 80\%. BitNest uses the same W4A8 representation for drafting while retaining
a W8A8 target, achieving an 89\% success rate. As shown in Table~\ref{tab:openvla_latency},
BitNest also increases decode throughput from 44.2 tok/s with FP16 AR
to 59.5 tok/s, corresponding to a $1.35\times$ decode-phase speedup. Overall, these results show that BitNest can accelerate autoregressive
action decoding without the task-level degradation observed with
standalone W4A8.

\begin{table}[th]
\centering
\caption{
Per-action latency breakdown on OpenVLA-7B. Numbers are median latencies over 100 frames on an NVIDIA A6000. Prefill includes the 288-token multimodal prefix and the first action token; decode generates the remaining six action tokens. Visual encoding and projection add 9.9\,ms equally to all methods and are excluded.
}
\label{tab:openvla_latency}

\small
\setlength{\tabcolsep}{6pt}
\renewcommand{\arraystretch}{1.06}

\begin{tabular}{lcccc}
\toprule
Method
& Prefill (ms)
& Decode (ms)
& Action Total (ms)
& Decode tok/s \\
\midrule
FP16 AR
& 65.9
& 135.6
& 201.5
& 44.2 \\

BitNest W8A8 AR
& 111.7
& 105.3
& 216.9
& 57.0 ($1.29\times$) \\

BitNest Spec. ($\gamma=7$)
& 111.7
& 100.8
& 212.7
& 59.5 ($1.35\times$) \\
\bottomrule
\end{tabular}
\end{table}
\paragraph{Latency breakdown.}
As shown in Table~\ref{tab:openvla_latency}, BitNest reduces decode
latency from 135.6\,ms with FP16 AR to 100.8\,ms, corresponding to a
$1.35\times$ decode-throughput speedup. Our INT8 Triton kernels are
optimized for small-$M$ decoding and verification, where $M$ denotes
the number of tokens processed by each linear layer and execution is
primarily bandwidth-bound. In this regime, directly reading the packed
4/8-bit weight planes reduces memory traffic.
OpenVLA prefill has $M=288$ and operates in a more compute-bound regime.
Our current large-$M$ path therefore dequantizes the packed weights to
FP16 and uses cuBLAS FP16 GEMM, while retaining activation quantization
and the other transformations used by the decoding path. These extra
operations make prefill about $1.7\times$ slower than FP16 in the
current implementation. Consequently, despite the 1.35$\times$ decode-phase speedup, our current implementation has a slightly higher per-action latency than FP16 AR (212.7 vs. 201.5 ms), as the prefill overhead outweighs the decode-time savings. Reducing this large-\(M\) prefill overhead, for example with a dedicated INT8 kernel, is left to future work.

\subsection{Additional Ablation Studies}
\label{app:ablation}
This section provides the detailed ablation results summarized in Section~\ref{sec:ablation}.
\subsubsection{Draft--Target Agreement}
\label{app:acceptance}

Table~\ref{tab:acceptance_rate} reports generation-time draft
accept rate across the evaluated workloads. BitNest maintains high
draft--target agreement despite using a W4A8 base for drafting and a
W8A8 refined representation for verification. Across all four models
and six workloads, BitNest achieves an average acceptance rate of
95.2\%, with model-wise averages ranging from 94.8\% to 95.7\%.

\begin{table*}[th]
\centering
\caption{
Generation-time draft acceptance rate (\%) across workloads on an
NVIDIA RTX A6000 with 2K prefixes, batch size 1, and 256 generated
tokens. Results use each method's native speculation setting.
Acceptance follows the native definition used by each method.
}
\label{tab:acceptance_rate}
\small
\setlength{\tabcolsep}{4.2pt}
\begin{tabular}{llcccccccc}
\toprule
Model & Method & $\gamma$
& WikiText-2 & GSM8K & Code & ShareGPT & LongDoc & PG19 & Avg. \\
\midrule

\multirow{4}{*}{LLaMA-2-7B}
& QSpec & 3
& 95.4 & 97.4 & 97.9 & 98.1 & 97.0 & 94.5 & 96.7 \\
& QuantSpec & 6
& 87.9 & 89.6 & 93.5 & 95.1 & 90.2 & 81.5 & 89.6 \\
& Draft \& Verify & Adaptive
& 62.6 & 82.7 & 82.8 & 84.3 & 83.0 & 72.2 & 77.9 \\
& \textbf{BitNest} & 4
& 92.5 & 95.9 & 96.9 & 97.8 & 96.7 & 94.2 & \textbf{95.7} \\
\midrule

\multirow{4}{*}{LLaMA-3-8B}
& QSpec & 3
& 92.0 & 97.0 & 97.1 & 97.5 & 94.5 & 90.8 & 94.8 \\
& QuantSpec & 6
& 77.5 & 88.9 & 87.2 & 94.6 & 80.9 & 77.7 & 84.5 \\
& Draft \& Verify & Adaptive
& 69.4 & 81.9 & 81.3 & 83.9 & 78.2 & 75.4 & 78.4 \\
& \textbf{BitNest} & 3
& 92.9 & 96.1 & 97.9 & 97.8 & 95.3 & 92.0 & \textbf{95.3} \\
\midrule

Qwen2-7B
& \textbf{BitNest} & 3
& 93.6 & 93.3 & 96.7 & 96.2 & 95.9 & 94.1 & \textbf{95.0} \\

Qwen2.5-7B
& \textbf{BitNest} & 3
& 92.1 & 95.4 & 95.0 & 96.8 & 95.4 & 93.9 & \textbf{94.8} \\
\midrule

\multicolumn{9}{r}{BitNest overall average}
& \textbf{95.2} \\
\bottomrule
\end{tabular}
\end{table*}

QSpec and QuantSpec maintain high agreement when the draft remains
close to the target representation. In contrast, BitNest retains
approximately 95\% acceptance while using the W4A8 base as the draft
and the nested W8A8 representation as the target. This indicates that
the base-first construction preserves strong alignment between the two
precision views, which is important for obtaining speculative speedup
from a substantially cheaper draft.

\subsubsection{Effect of Speculation Length}
\label{app:spec_length}
\begin{table}[th]
\centering
\caption{
Effect of speculation length $\gamma$ on LLaMA-2-7B-32K with KV4
drafting. Each entry reports decode throughput in tok/s, with the
draft-slot acceptance rate in parentheses.
}
\label{tab:gamma_ablation}
\small
\setlength{\tabcolsep}{5pt}

\begin{tabular}{lccccc}
\toprule
Context & $\gamma=2$ & $\gamma=3$ & $\gamma=4$ & $\gamma=5$ & $\gamma=6$ \\
\midrule
4K  & 57.2 (.913) & 56.8 (.897) & \textbf{59.3 (.856)} & 56.1 (.820) & 55.6 (.793) \\
31K & 33.0 (.901) & 33.6 (.881) & \textbf{34.3 (.875)} & 34.0 (.834) & 33.0 (.797) \\
\bottomrule
\end{tabular}
\end{table}
As shown in Table~\ref{tab:gamma_ablation}, increasing $\gamma$ reduces the draft-slot acceptance rate, while throughput first improves and then decreases. In both the 4K and 31K settings, $\gamma=4$ provides the highest throughput, and we therefore use it as the default for the corresponding LLaMA experiments.

\subsubsection{Effect of Draft KV Precision}
\label{app:draft_kv}
To further explain the long-context behavior in
Figure~\ref{fig:long_context}(a), we compare the effect of draft KV
precision on models with different attention architectures. Table~
\ref{tab:draft_kv_precision} reports decode throughput at 32K context
while varying only the KV precision accessed during drafting.

\begin{table}[th]
\centering
\caption{
Effect of draft KV precision at 32K context across different attention
architectures. All entries report decode throughput in tok/s.
}
\label{tab:draft_kv_precision}
\small
\setlength{\tabcolsep}{6pt}
\begin{tabular}{llccc}
\toprule
Model & Attention & BF16 KV & KV8 & KV4 \\
\midrule
LLaMA-2-7B-32K & MHA & 23.4 & 28.5 & \textbf{34.3} \\
Qwen2.5-7B     & GQA & 50.7 & 51.1 & \textbf{51.4} \\
\bottomrule
\end{tabular}
\end{table}

As shown in Table~\ref{tab:draft_kv_precision}, draft KV precision has
a substantially larger effect on LLaMA-2-7B-32K than on Qwen2.5-7B.
LLaMA-2 uses MHA with 32 KV heads, so KV-cache traffic becomes a major
fraction of the decoding cost at long contexts. Reducing draft access
from BF16 KV to KV8 and further to KV4 increases throughput from 23.4
to 28.5 and 34.3 tok/s, respectively. Notably, KV4 drafting achieves
the highest throughput despite a lower acceptance rate than KV8
drafting (0.875 vs.\ 0.944), showing that reducing draft cost can
outweigh a moderate decrease in acceptance.

In contrast, Qwen2.5 uses GQA, which reduces KV-cache traffic by
sharing key and value heads across multiple query heads. At 32K
context, BF16, KV8, and KV4 drafting achieve 50.7, 51.1, and
51.4 tok/s, respectively. The small difference indicates that
reducing draft KV precision provides little additional throughput
benefit when KV-cache traffic is already substantially reduced by GQA. These results show that the latency benefit of low-precision draft KV
depends strongly on the underlying attention architecture and becomes
most pronounced when KV-cache traffic is a dominant decoding cost.

\subsubsection{Additional Target-Precision Analysis}
\label{app:target_precision}

This section provides the detailed target-precision ablations summarized
in Table~\ref{tab:target_precision_ablation} of the main text, covering
both KV-cache and weight precision.

\begin{table}[th]
\centering
\caption{
Detailed ablation on draft and target KV precision.
The target weights are fixed to W8A8 and speculative decoding uses
$\gamma=4$. GSM8K reports 8-shot accuracy on 500 examples.
For LongBench, Qasper, HotpotQA, and MultiFieldQA report F1;
SAMSum and GovReport report ROUGE-L; TREC reports accuracy.
}
\label{tab:app_kv_precision_full}

\scriptsize
\setlength{\tabcolsep}{2.7pt}
\renewcommand{\arraystretch}{1.06}

\resizebox{\linewidth}{!}{
\begin{tabular}{llccccccccc}
\toprule
Model & Configuration & GSM8K$\uparrow$ & Qasper$\uparrow$ & HotpotQA$\uparrow$ & MultiFieldQA$\uparrow$ & SAMSum$\uparrow$ & GovReport$\uparrow$ & TREC$\uparrow$ & Avg. & Accept$\uparrow$ \\
\midrule

\multirow{6}{*}{LLaMA-2-7B-32K}
& FP16 AR & -- & 34.6 & 47.9 & 39.8 & 36.8 & -- & 71.0 & 46.0 & -- \\
& W8A8+KV8 AR & -- & 34.1 & 47.9 & 40.3 & 36.8 & 20.8 & 69.5 & 45.7 & -- \\
& KV8 draft $\rightarrow$ KV8 target & -- & 33.9 & 47.9 & 40.3 & 37.1 & 20.7 & 69.5 & 45.7 & 0.970 \\
& KV4 draft $\rightarrow$ KV8 target & -- & 33.7 & 47.9 & 40.3 & 36.9 & 20.7 & 69.5 & 45.7 & 0.946 \\
& W8A8+KV4 AR & -- & 33.5 & 47.2 & 39.9 & 36.7 & 20.6 & 69.5 & 45.4 & -- \\
& KV4 draft $\rightarrow$ KV4 target & -- & 32.9 & 46.7 & 40.0 & 37.1 & 20.7 & 69.5 & 45.3 & 0.965 \\

\midrule

\multirow{5}{*}{Qwen2.5-7B}
& W8A8+KV8 AR & 82.4 & 45.1 & 32.4 & 35.6 & 41.0 & 21.6 & 72.5 & 41.4 & -- \\
& KV8 draft $\rightarrow$ KV8 target & 82.2 & 44.8 & 32.4 & 35.5 & 40.6 & 21.6 & 72.5 & 41.2 & 0.953 \\
& KV4 draft $\rightarrow$ KV8 target & 81.6 & 44.8 & 32.9 & 35.7 & 40.6 & 21.6 & 72.5 & 41.3 & 0.937 \\
& W8A8+KV4 AR & 80.2 & 45.0 & 31.9 & 37.3 & 40.5 & 22.0 & 72.5 & 41.5 & -- \\
& KV4 draft $\rightarrow$ KV4 target & 78.4 & 44.8 & 31.9 & 37.1 & 40.4 & 21.7 & 72.5 & 41.4 & 0.933 \\

\bottomrule
\end{tabular}
}

\vspace{2pt}
{\scriptsize
For LLaMA-2-7B-32K, Avg. is computed over the five tasks shared with
the FP16 reference and excludes GovReport. For Qwen2.5-7B, Avg. is
computed over all six LongBench tasks.
}
\end{table}

\paragraph{Target KV precision.}
Table~\ref{tab:app_kv_precision_full} provides the full target-KV ablation.
On Qwen2.5-7B, reducing the target KV cache from KV8 to KV4 decreases
GSM8K accuracy from 82.4\% to 80.2\% under autoregressive decoding,
while using KV4 only for drafting retains 81.6\%. When both drafting
and verification use KV4, accuracy further decreases to 78.4\%.
In contrast, the average LongBench score remains relatively stable
across these configurations, indicating that sensitivity to target KV
precision is task-dependent.

A similar trend is observed on LLaMA-2-7B-32K. Using KV4 only during
drafting preserves the average LongBench score of the KV8 target
(45.7), whereas using KV4 for both drafting and verification reduces
it to 45.3. These results support using lower-precision KV access on
the speculative path while retaining KV8 for final verification.

\begin{table}[th]
\centering
\caption{
Detailed ablation on target weight precision.
For language models, each entry reports WikiText-2 perplexity
($\downarrow$) / GSM8K 5-shot strict accuracy ($\uparrow$).
W4A16 isolates the effect of 4-bit weights from activation quantization.
}
\label{tab:app_weight_precision}

\scriptsize
\setlength{\tabcolsep}{3.5pt}
\renewcommand{\arraystretch}{1.08}

\resizebox{\linewidth}{!}{
\begin{tabular}{lcccccc}
\toprule
Model & FP16 & W8A8 & W4A16 & W4A8 & BitNest target & W4A4 \\
\midrule
LLaMA-2-7B & 5.471 / 13.6\% & 5.476 / 13.4\% & 5.599 / 11.9\% & 5.591 / 12.2\% & 5.512 / 13.8\% & 6.717 / 5.8\% \\
LLaMA-3-8B & 6.139/ 50.1\% & 6.162 / 50.0\% & 6.470 / 41.8\% & 6.476 / 41.4\% & 6.223 / 49.8\% & 8.660 / 26.6\% \\
Qwen2-7B & 7.134 / 79.4\% & 7.152 / 78.9\% & 7.238 / 75.5\% & 7.244 / 75.5\% & 7.176 / 77.6\% & 8.079 / 63.3\% \\
Qwen2.5-7B & 6.851 / 82.9\% & 6.872 / 83.2\% & 6.998 / 80.1\% & 7.017 / 79.3\% & 6.906 / 82.0\% & 8.085 / 66.7\% \\
\bottomrule
\end{tabular}
}
\end{table}
\paragraph{Target weight precision.}
Table~\ref{tab:app_weight_precision} provides a detailed comparison across
target weight precisions. Directly using the W4 base as the final model
can noticeably degrade generation quality, even when activation precision
is increased. For example, on LLaMA-3-8B, W4A8 and W4A16 achieve
41.4\% and 41.8\% GSM8K accuracy, respectively, compared with 50.0\%
for W8A8. In contrast, the BitNest target recovers to 49.8\% while
preserving the same W4 base used for drafting.

The same trend is observed across the other evaluated models. On
Qwen2.5-7B, W4A8 achieves 79.3\% GSM8K accuracy, whereas the BitNest
target reaches 82.0\%, close to the 83.2\% W8A8 baseline. These results
show that the low-precision base is suitable for speculative drafting,
while the refinement is important for recovering the fidelity required
for final prediction.

\subsection{Additional Edge-Device Results}
\label{app:edge}

This section provides additional edge-device results complementing
Section~\ref{sec:edge}. We evaluate BitNest on NVIDIA Jetson Orin NX
devices with 16GB and 8GB memory. Experiments use batch size 1 and
greedy decoding, and throughput is measured over the decode phase.

\subsubsection{Additional Model Results}

Table~\ref{tab:edge_all_models} reports decoding throughput across the
evaluated 7B and 3B models. On Orin NX16, BitNest reaches
9.0--9.3 tok/s on LLaMA-2-7B-32K, compared with 7.8 tok/s for W8A8
AR and 6.0 tok/s for FP16 AR. On Qwen2.5-7B, FP16 inference exceeds
the available device memory, while BitNest reaches 8.37--8.50 tok/s
compared with 7.37 tok/s for W8A8 AR.

The same trend holds on the 8GB device. BitNest reaches
15.5--16.6 tok/s on Llama-3.2-3B and 15.2 tok/s on Qwen2.5-3B,
compared with 14.0 and 14.1 tok/s for W8A8 AR, respectively.

\begin{table}[th]
\centering
\caption{
Edge-device decoding performance across the evaluated models.
Throughput is measured over the decode phase. OOM denotes configurations
that exceed the available device memory.
}
\label{tab:edge_all_models}
\small
\setlength{\tabcolsep}{5.5pt}
\begin{tabular}{llccc}
\toprule
Model & Device & FP16 AR & W8A8 AR & BitNest \\
\midrule
LLaMA-2-7B-32K
& NX16
& 6.0
& 7.8
& \textbf{9.0--9.3} \\

Qwen2.5-7B
& NX16
& OOM
& 7.37
& \textbf{8.37--8.50} \\

Llama-3.2-3B
& NX8
& 10.6
& 14.0
& \textbf{15.5--16.6} \\

Qwen2.5-3B
& NX8
& 11.0
& 14.1
& \textbf{15.2} \\
\bottomrule
\end{tabular}

\vspace{2pt}
{\footnotesize All throughput values are in tok/s.}
\end{table}

\subsubsection{Memory Footprint}

Table~\ref{tab:edge_memory} reports peak memory at 2K context together
with the maximum executable context length on Orin NX16. On
LLaMA-2-7B-32K, BitNest reduces PyTorch peak allocation from
13.9 GiB with FP16 AR to 7.7 GiB, while maintaining a memory footprint
comparable to W8A8 AR. The reduced footprint also extends the maximum
executable context from 2K with FP16 AR and 6K with W8A8 AR to 10K
with BitNest. On Qwen2.5-7B, FP16 inference exceeds the available
memory at 2K context, whereas both W8A8 AR and BitNest remain
executable. Overall, BitNest supports speculative decoding without
storing a separate draft-weight representation while retaining the
memory efficiency of quantized inference.

\begin{table}[th]
\centering
\caption{
Edge-device memory usage and context capacity on Orin NX16.
Peak memory is measured at 2K context and reported as
PyTorch peak allocation / system-wide peak memory in GiB.
}
\label{tab:edge_memory}
\small
\setlength{\tabcolsep}{6pt}
\begin{tabular}{llcc}
\toprule
Model & Method & Peak Memory (GiB) & Max. Context \\
\midrule
\multirow{3}{*}{LLaMA-2-7B-32K}
& FP16 AR & 13.9 / 15.4 & 2K \\
& W8A8 AR & 8.2 / 11.4 & 6K \\
& BitNest & \textbf{7.7 / 10.8} & \textbf{10K} \\
\midrule
\multirow{3}{*}{Qwen2.5-7B}
& FP16 AR & OOM & -- \\
& W8A8 AR & 9.0 / 13.7 & -- \\
& BitNest & \textbf{8.9 / 13.5} & -- \\
\bottomrule
\end{tabular}
\end{table}

\subsubsection{Effect of GEMV Kernel Optimization}

The edge-device results reported above use our optimized GEMV kernel.
The original low-bit kernel was designed for the bandwidth-bound
small-$M$ regime but reached only about 44\% of the estimated
memory-bandwidth roofline. Split-K tuning alone provided negligible
improvement, motivating a redesigned GEMV implementation.

On LLaMA-2-7B-32K at 2K context, the optimized kernel increases W8A8
AR throughput from approximately 6.0 to 7.8 tok/s. BitNest improves
from 7.4--7.9 tok/s with the original implementation to
9.0--9.3 tok/s. Correspondingly, the effective bandwidth of the W8
kernel increases from approximately 46 to 63 GB/s. These results show
that efficiently exploiting the reduced weight traffic is important for
realizing the benefit of the nested representation on bandwidth-limited
edge hardware.

\end{document}